\documentclass[conference]{IEEEtran}
\IEEEoverridecommandlockouts
\usepackage{cite}
\usepackage{amsmath,amssymb,amsfonts}
\usepackage{algorithmic}
\usepackage{lipsum,adjustbox}
\usepackage{verbatim}
\usepackage{makecell}
\usepackage{graphics}
\usepackage{stackengine}
\usepackage{pgfplots}
\pgfplotsset{width=7cm,compat=1.8}
\usepackage{multirow}
\usepackage{soul}
\usepackage{amssymb}
\usepackage{amsmath}
\usepackage[caption=false,font=footnotesize]{subfig}
\usepackage{algorithm}
\usepackage{algorithmic}
\usepackage{pgfplots}
\pgfplotsset{compat=1.18}
\usepackage{graphicx}
\usepackage{textcomp} 
\usepackage{booktabs}
\usepackage{stfloats}
\usepackage{scalefnt}
\usepackage{mathrsfs}       
\DeclareMathAlphabet{\pazocal}{OMS}{zplm}{m}{n}
\usepackage[nolist]{acronym}
    
 \usepackage{bm}

\newcommand{\compl}{\mathbb{C}}         

\newcommand{\ma}  [1]{ \bm{#1} } 

\newcommand{\set} [1]{{\mathcal {#1}}} 
\newcommand{\Kon} {\set{K}_{\text{on}}} 

\begin{acronym}
    \acro{C-ITS}{cooperative intelligent transportation system}
    \acro{RBF}{radial basis function}
    \acro{ANN}{artificial neural network}
\acro{GRU}{gated recurrent unit}
\acro{RNN}{Recurrent neural network}
\acro{Bi-RNN}{bi-directional recurrent neural network}
\acro{Bi}{bi-directional}
\acro{FNN}{Feed-forward Neural Network}
    \acro{CNN}{convolutional neural network}
    \acro{TS-ChannelNet}{Temporal spectral ChannelNet}
    \acro{TDR}{transmission data rate}
    \acro{LSTM}{long short-term memory}
    \acro{ALS}{Accurate LS}
    \acro{SLS}{simple LS}
    \acro{LRP}{Layer-wise Relevance Propagation}
    \acro{XAI}{explainable artificial intelligence}
    \acro{ChannelNet}{channel network}
    \acro{ADD-TT}{average decision-directed with time truncation}
    \acro{WI}{weighted interpolation}
    \acro{DD}{decision-directed}
    \acro{SR-ConvLSTM}{super-resolution convolutional long short-term memory}
    \acro{SR-CNN}{super resolution CNN}
    \acro{DN-CNN}{denoising CNN}
    \acro{V2V}{vehicle-to-vehicle}
    \acro{V2I}{vehicle-to-infrastructure}
    \acro{VTV}{vehicle-to-vehicle}
    \acro{RTV}{roadside-to-vehicle}
    \acro{DPA}{data-pilot aided}
    \acro{STA}{spectral temporal averaging}
    \acro{M-STA}{modified STA}
    \acro{LMMSE}{linear minimum mean-squared error}
    \acro{M-CDP}{modified CDP}
    \acro{M-TRFI}{modified TRFI}
    \acro{SNR}{signal-to-noise ratio}
    \acro{CDP}{Constructed data pilots}
    \acro{DFT}{discrete Fourier transform}
    \acro{T-DFT}{truncated discrete Fourier transform}
    \acro{TA-MDFT}{temporal averaging M-DFT}
    \acro{TA-DFT}{TA-DFT}
    \acro{STS}{short training symbols}
    \acro{LTS}{long training symbols}
    \acro{SS}{signal symbol}
    \acro{TDL}{tapped delay line}
    \acro{TRFI}{time domain reliable test frequency domain interpolation}
    \acro{SBS}{symbol-by-symbol}
    \acro{FBF}{frame-by-frame}
    \acro{MMSE-VP}{minimum mean square error using virtual pilots}
    \acro{DL}{deep learning}
    \acro{AE-DNN}{auto-encoder deep neural network}
    \acro{DNN}{deep neural networks} 
    \acro{ISI}{inter-symbol-interference}
    \acro{OFDM}{Orthogonal Frequency-Division Multiplexing}
    \acro{LS}{least square}
    \acro{RS}{reliable subcarriers}
    \acro{URS}{unreliable subcarriers}
    \acro{SRNN}{Simple RNN}
    \acro{MMSE}{minimum mean squared error}
    \acro{SoA}{state of the art}
    \acro{NMSE}{normalized mean-squared error}
    \acro{AWGN}{additive white Gaussian noise}
	\acro{BER}{Bit Error Rate}
	\acro{RSU}{road side unit}
	\acro{AE}{auto-encoder}
    \acro{CP}{cyclic prefix}
\acro{PLCP}{physical layer convergence protocol}
\acro{MSE}{mean squared error}
\end{acronym}

\usepackage[utf8]{inputenc}
\usepackage[english]{babel}
\usepackage{amsthm}
\usepackage{commath}

\newcommand{\Lb}{\pazocal{L}}

\usepackage{cleveref}
\crefname{lemma}{Lemma}{Lemmas}
\usepackage{flushend}

\begin{document}

\title{X-REFINE: XAI-based RElevance input-Filtering and archItecture fiNe-tuning for channel Estimation
\thanks{This work was supported by the framework of the Travel Project through French National Research Agency (ANR) under Grant ANR-24-IAS1-0003.}
}

\author{\IEEEauthorblockA{Abdul Karim Gizzini\IEEEauthorrefmark{1},
Yahia Medjahdi\IEEEauthorrefmark{4}
}

\IEEEauthorblockA{\IEEEauthorrefmark{1}University of Paris-Est Créteil (UPEC), LISSI/TincNET, F-94400, Vitry-sur-Seine, France.\\
\IEEEauthorrefmark{4} IMT Nord Europe, Institut Mines T\'el\'ecom, Centre for Digital Systems, F-59653 Villeneuve d’Ascq, France. \\
Email: abdul-karim.gizzini@u-pec.fr, yahia.medjahdi@imt-nord-europe.fr}}



\maketitle

\begin{abstract}
AI-native architectures are vital for 6G wireless communications. The black-box nature and high complexity of deep learning models employed in critical applications, such as channel estimation, limit their practical deployment. While perturbation-based eXplainable Artificial Intelligence (XAI) solutions offer input filtering, they often neglect internal structural optimization. We propose X-REFINE, an XAI-based framework for joint input-filtering and architecture fine-tuning. By utilizing a decomposition-based, sign-stabilized LRP-$\epsilon$ rule, X-REFINE backpropagates predictions to derive high-resolution relevance scores for both subcarriers and hidden neurons. This enables a reliable optimization that identifies the most reliable model components. Simulation results demonstrate that X-REFINE achieves a superior performance-complexity-interpretability trade-off compared to the external perturbation-based XAI frameworks, significantly reducing computational complexity while maintaining robust bit error rate (BER) performance.
\end{abstract}


\section{Introduction} \label{introduction}

The shift toward AI-native architectures in 6G wireless systems~\cite{11435025, 11217271} is driven by the need for autonomous, self-optimizing systems capable of meeting Ultra-Reliable Low-Latency Communication (URLLC) requirements~\cite{11126933}. In V2X communications, Deep Learning (DL)-based denoising models, such as \acp{FNN}, have significantly improved channel estimation over classical schemes. However, these models operate as black boxes lacking the transparency required for critical applications where reliability, safety, and trust must be guaranteed~\cite{10742571}. The black-box nature hides the model's internal decision methodology, making it hard to interpret how it processes the inputs. Consequently, determining the model's input features and internal architecture often relies on empirical simulations.

Recent research has introduced eXplainable AI (XAI) as a solution that provides interpretability of the black-box DL models by identifying which input features contribute more to the final prediction~\cite{10854503,10620685}. In~\cite{gizzini2025explainable} we proposed a perturbation-based XAI framework for channel estimation, denoted as XAI-CHEST. XAI-CHEST employs an external noise model that induces noise on top of a pretrained model's inputs without degrading its performance. Hence, it offers a learnable input filtering strategy where only the relevant inputs can be used to improve the overall performance. Perturbation-based XAI schemes rely on external processing, deriving relevance scores without considering the model's internal structural parameters. These methods are constrained to input-level optimization, failing to address the architectural design. However, further architecture pruning provides better performance-complexity-interpretability trade-offs.

In this context, we propose in this paper an \textbf{X}AI-based \textbf{RE}levance input-\textbf{F}iltering and arch\textbf{I}tecture fi\textbf{N}e-tuning for channel \textbf{E}stimation framework called X-REFINE. The main objective is to jointly optimize both the model inputs and architecture, resulting in a fully optimized model. Unlike perturbation-based XAI schemes, X-REFINE leverages the decomposition-based sign-stabilized \ac{LRP}-$\epsilon$ rule~\cite{montavon2019layer} to identify relevant model components. The relevance scores are derived by backpropagating the model prediction to the model inputs. This allows X-REFINE to derive relevance scores with better certainty than external perturbation-based schemes. Therefore, better interpretability resolution is achieved via faithful relevant input selection and the optimization of the model architecture. Simulation results demonstrate that X-REFINE can significantly reduce the overall computational complexity by jointly optimizing the input feature set and the internal architecture, while maintaining a robust \ac{BER} performance and system efficiency. To sum up, the contributions of this work can be summarized as follows:


\begin{itemize}
    \item Proposing a decomposition-based XAI scheme, denoted as X-REFINE, which leverages the sign-stabilized LRP-$\epsilon$ rule to identify, simultaneously, relevant model components, i.e, inputs and architecture.
    \item Deriving the analytical expression and the corresponding simulations of both the model inputs and architecture optimization.

    \item Showing that using the relevant inputs and neurons instead of the full sets maintains a robust performance while significantly reducing the required computational complexity.
\end{itemize} 

The remainder of this paper is organized as follows: Section II presents the system model. The proposed X-REFINE double optimization scheme is presented in Section III. In Section IV, the performance evaluation and the computational complexity are analyzed. Finally, Section V concludes the paper.

\textbf{Notations}: Throughout the paper, vectors and scalars are defined by lowercase bold and uppercase symbols, respectively.

\section{System Model} \label{system_model}

In this work, we consider an {\ac{OFDM}} system. An initial channel estimation is performed prior to the {\ac{FNN}} processing. This acts as a denoising unit to improve the initially estimated channel.  Let ${\ma{y}}_{{q}} \in \compl^{K_{\text{on}} \times 1}$ be the $q$-th received frequency-domain {\ac{OFDM}} symbol expressed as:
\begin{equation}
	\begin{split}
		{\ma{y}}_{{q}}[k] 
		&= {\ma{h}}_q[k] {\ma{x}}_q[k] +  {\ma{e}}_{q}[k] + {\ma{v}}_q[k],~ k \in \Kon.
	\end{split}            
	\label{eq: system_model}
\end{equation}

${\ma{x}}_q \in \compl^{K_{\text{on}} \times 1}$ is the $q$-th transmitted frequency-domain {\ac{OFDM}} symbol which includes $ {\ma{x}}_{{q,d}} \in \compl^{K_{d} \times 1}$ data subcarriers, $ {\ma{x}}_{{q,p}} \in \compl^{K_{p} \times 1}$ pilot subcarriers, and $K_{n}$ guard band subcarriers. We note that $K_{\text{on}} = K_{p} + K_{d}$ represents the active subcarriers out of $K$ total subcarriers. ${\ma{h}}_q \in \compl^{K_{\text{on}} \times 1}$ refers to the frequency response of the doubly-selective channel and ${{\ma{v}}}_q$ represents the \ac{AWGN} noise of variance $\sigma^2$. ${\ma{e}}_{q}$ denotes the Doppler-induced inter-carrier interference derived in~\cite{9743925}.

{\ac{FNN}}-based channel estimation is considered, which employs conventional estimation denoted by $\hat{{\ma{h}}}_{\Phi_{q}}$ prior to the FNN processing, such that:

\begin{equation}
    \hat{{\ma{h}}}_{\Phi\text{-FNN}_{q}} = \mathcal{F}(\hat{{\ma{h}}}_{\Phi_{q}}, \Theta),
    \label{eq: STA_FNN}
\end{equation}
where $\Theta = \{W^{(l)}\}_{l=1}^L$ represents the weights of the employed FNN architecture with size $|\Theta|$ and $L$ denotes the number of the FNN layers. $\Phi$ is the employed initial channel estimation scheme and $\mathcal{F}(.)$ denotes the unoptimized FNN model with full input and full architecture.
\section{Proposed X-REFINE Framework} \label{Proposed_scheme}

This section presents the proposed X-REFINE double optimization framework, followed by the LRP theoretical aspects and the computational complexity reduction offered by the X-REFINE framework.

\subsection{X-REFINE Double Optimization Problem}

The proposed X-REFINE framework is modeled as a double optimization problem that simultaneously filters the relevant input features and prunes the employed FNN model architecture. The objective is to determine the optimal binary decision input mask $\ma{m}^{*}_{\text{in}} \in \{0,1\}^{K{\text{on}}}$ and architectural mask $\ma{m}^{*}_{\text{arch}} \in \{0,1\}^{|\Theta|}$ that minimize the \ac{MSE} of the optimized FNN model with relevant inputs and pruned architecture denoted as $\mathcal{F}^{*}(.)$. Hence, the optimized FNN-based channel estimate $\hat{{\ma{h}}}^{*}_{\Phi\text{-FNN}_{q}}$ can be expressed as:

\begin{equation}
    \hat{{\ma{h}}}^{*}_{\Phi\text{-FNN}_{q}} = \mathcal{F}^{*} \left( \underbrace{\hat{{\ma{h}}}_{\Phi_{q}} \odot \ma{m}^{*}_{\text{in}}}_{\text{Input Filtering}} ; \underbrace{\Theta \odot \ma{m}^{*}_{\text{arch}}}_{\text{Model Pruning}} \right).
    \label{eq:fnnopt}
\end{equation}

$\odot$ denotes the element-wise Hadamard product. X-REFINE double optimization problem aims to minimize the Mean Squared Error (MSE) between the true channel, $\ma{h}_{i}$, and the optimized FNN-based channel estimate defined in~{\eqref{eq:fnnopt}}, such that:
\begin{equation}
\begin{aligned}
\min_{\tau, P} \quad &  \Lb_{\mathcal{F}^{*}} = \frac{1}{N_{\text{tr}}}. \sum_{i = 1}^{N_{tr}} \left( {\ma{h}}_i - \mathcal{F}^{*} \Big( \hat{{\ma{h}}}^{*}_{\Phi_{i}} \odot \ma{m}^{*}_{\text{in}} (\tau) ; \Theta \odot \ma{m}^{*}_{\text{arch}}(P)\Big)\right)^{2},\\
\textrm{s.t.} \quad &  \ma{m}^{*}_{\text{in}} = \mathbb{I} \left( \bar{\ma{r}}_{\text{in}} \geq \tau \right), \quad \tau \in \mathcal{T}, \\
& \ma{m}^{*}_{\text{arch}} = \mathbb{I} \left( \bar{\ma{r}}_{\text{arch}} \geq \text{Percentile}(\bar{\ma{r}}^{(l)}, P) \right), P \in \mathcal{P}, \\
& \text{BER}(\mathcal{F}^{*}) < \text{BER}(\mathcal{F}).
\end{aligned}
\label{eq:threshold_fine_tuning}
\end{equation}


$N_{\text{tr}}$ denotes the number of training samples. $\mathbb{I}(\cdot)$ is the indicator function, which equals 1 if the corresponding condition is met and 0 otherwise. $\bar{\ma{r}}_{\text{in}}$ and $\bar{\ma{r}}_{\text{arch}}$ are the global relevance scores assigned to the FNN inputs and the $l$-th hidden layer, respectively. $N_{tr}$ denotes the number of training samples. We note that $\tau$ and $P$ refer to the optimal input relevance threshold and the optimal architectural pruning percentile (in $\%$), where $\mathcal{T} = \{ \tau \in \mathbb{R} \mid \tau_{\text{min}} \leq \tau \leq \tau_{\text{max}}, \tau = \tau_{\text{min}} + n_{\tau} \Delta\tau\}$, and $\mathcal{P} = \{ P \in \mathbb{Z} \mid 0 \leq P \leq P_{\text{max}}, P = n_P \Delta P\}$. $\Delta\tau$ and $\Delta P$ refer to the step sizes with $ n_{\tau}$ and $n_{P}$ step indices, respectively. Finally, the average {\ac{BER}} for the training SNR employing $\mathcal{F}^{*}$, must be better than that employing $\mathcal{F}$.

\begin{figure*}[t]
\centering
\includegraphics[width=2\columnwidth, height=6cm]{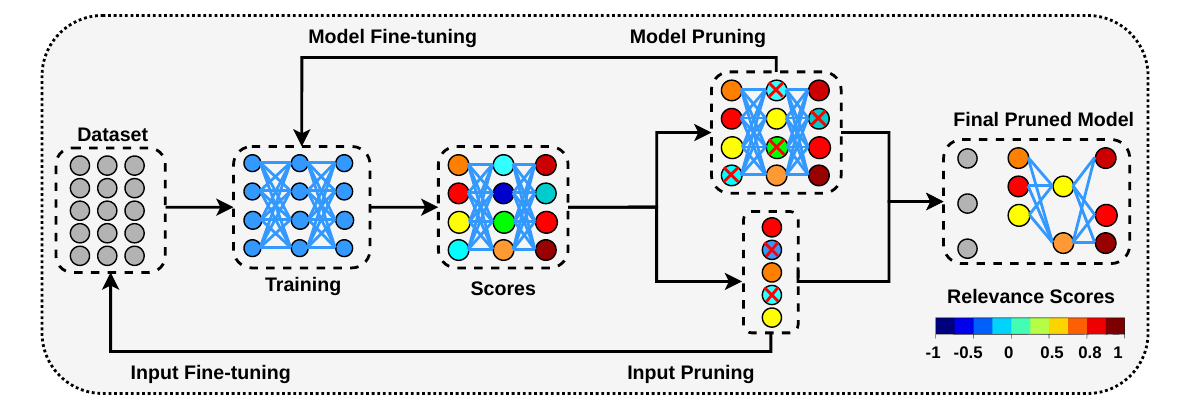}
\caption{Block diagram of the proposed X-REFINE framework.
\label{fig:proposed_lrp_scheme}}
\end{figure*}

X-REFINE jointly optimizes $\ma{m}^{*}_{\text{in}}$ and $\ma{m}^{*}_{\text{arch}}$ by leveraging XAI-derived relevance scores. Input filtering employs value-based thresholding ($\bar{\ma{r}}_{i} \geq \tau$) to exploit channel sparsity, dynamically excluding subcarriers that fall below a fixed quality-control threshold. Conversely, percentile-based thresholding ($P$) is applied to hidden layers to enforce structural sparsity. By pruning a fixed proportion of neurons (e.g., the bottom 30\%), we guarantee model compression and prevent layer collapse, as relevance scores naturally scale down with depth. While input filtering removes irrelevant subcarriers, percentile ranking maintains network integrity by preserving the most significant features relative to each layer's specific distribution. The optimization problem in (4) involves the indicator function $\mathbb{I}(\cdot)$, rendering the objective function non-differentiable and the search space combinatorial. Since the mapping between the discrete masks $(\ma{m}^{*}_{\text{in}}, \ma{m}^{*}_{\text{arch}})$ and the BER performance is non-convex and lacks a closed-form derivative, standard gradient-based optimization techniques are inapplicable. Consequently, we adopt a structured numerical grid search over the discrete sets $\mathcal{T}$ and $\mathcal{P}$. This approach provides a computationally tractable way to identify the global optimum within the defined search space while empirically guaranteeing that the BER constraints are satisfied.

\subsection{LRP XAI Scheme}

X-REFINE employs the Sign-Stabilized LRP-$\epsilon$ rule to compute relevance scores for model inputs and architecture. LRP decomposes the output $\hat{{\ma{h}}}_{\Phi\text{-FNN}_{q}}$ via a conservation law, ensuring total relevance at any layer $l$ is preserved:

\begin{equation}
\sum_{i} \bar{\ma{r}}_i^{(l)} = \sum_{j} \bar{\ma{r}}_j^{(l+1)} = |\hat{{\ma{h}}}_{\Phi\text{-FNN}{q}}|.\end{equation}

To address polarity-induced cancellation in complex-valued regression, we initialize the relevance as the norm of $|\hat{{\ma{h}}}_{\Phi\text{-FNN}_{q}}|$. This phase-invariant initialization ensures magnitude-based importance mapping, preventing negative scores of the FNN's phase offsets from distorting the attribution. Consequently, X-REFINE robustly identifies subcarrier contributions, where negative relevances solely reflect learned weights rather than output signs. To filter noise and enhance stability, the relevance of a neuron $i$ in layer $l$, denoted as $\bar{\ma{r}}_i^{(l)}$\footnote{The relevance scores for the input layer, $\bar{\ma{r}}_i^{(0)}$, are computed for the $2K_{\text{on}}$ inputs, representing the separate real and imaginary components of the channel estimates. These scores are then averaged for each subcarrier, resulting in $\bar{\ma{r}}_\text{in} \in \{0, 1\}^{K_{\text{on}}}$ and the optimal input mask $\ma{m}^{*}_{\text{in}} \in \{0, 1\}^{K_{\text{on}}}$ = $\mathbb{I} (\bar{\ma{r}}_\text{in})$.}, is computed by aggregating the relevance scores from all neurons in the subsequent layer $l+1$, such that: 

\begin{equation}
\bar{\ma{r}}_i^{(l)} = \sum_j \left( \frac{z_{ij}}{\sum_k z_{kj} + \epsilon \cdot \text{sign}(\sum_k z_{kj})} \right) \bar{\ma{r}}_j^{(l+1)},
\label{eq6}
\end{equation}
where $z_{ij} = a_i^{(l)} w_{ij}^{(l, l+1)}$ represents the forward-pass contribution of neuron $i$ to the activation of neuron $j$, consisting of the activation $a_i^{(l)}$ and the interconnecting weight $w_{ij}^{(l, l+1)}$. $\sum_{k} z_{kj}$ is the total input to neuron $j$ from the preceding layer. $\epsilon$ is a small sign-dependent stabilizer used to prevent numerical collapse when $\sum_k z_{kj} \approx 0$ without altering attribution polarity. Normalizing each connection strength $z_{ij}$ by the total successor input ensures that the redistribution respects the contribution's original sign while maintaining structural stability. Algorithm~\ref{alg:x_refine_iterative} summarizes the proposed X-REFINE framework.

\begin{algorithm}[t]
\caption{Iterative X-REFINE Joint Optimization. $\text{Val-Loss}_{\text{p}}$ and $\text{Val-Loss}_{\text{c}}$ denote the previous and current validation losses, respectively.}
\label{alg:x_refine_iterative}
\begin{algorithmic}[1]
\REQUIRE Trained $\mathcal{F}$, dataset $\hat{{\ma{H}}}_{\Phi}$, $\mathcal{T}, \mathcal{P}$, $\text{BER}_{\text{target}}$
\ENSURE $\ma{m}^{*}_{\text{in}}, \ma{m}^{*}_{\text{arch}}$.
\STATE \textbf{Phase I: Relevance Mapping}
\STATE Compute global averages $\bar{\ma{r}}_{\text{in}}, \bar{\ma{r}}_{\text{arch}}$ via LRP-$\epsilon$ rule over $N$ samples.
\STATE \textbf{Phase II: Iterative Pruning}
\STATE Initialize $\text{Val-Loss}_{\text{p}}$
\FORALL{$\tau \in \mathcal{T}, P \in \mathcal{P}$}
    \STATE $\ma{m}_{\text{in}} \gets (\bar{\ma{r}}_{\text{in}} \geq \tau)$, $\ma{m}_{\text{arch}} \gets ((\bar{\ma{r}}_{\text{arch}}) > P)$
    \STATE \textbf{Retrain} $\mathcal{F}$ with $(\ma{m}_{\text{in}}, \ma{m}_{\text{arch}})$ $\rightarrow$ Obtain $\text{Val-Loss}_{\text{c}}$
    \IF{$\text{Val-Loss}_{\text{c}} < \text{Val-Loss}_{\text{p}}$ \AND $\text{BER} \leq \text{BER}_{\text{target}}$}
        \STATE $\ma{m}^{*}_{\text{in}} \gets \ma{m}_{\text{in}}$, $\ma{m}^{*}_{\text{arch}} \gets \ma{m}_{\text{arch}}$
        \STATE $\text{Val-Loss}_{\text{p}} \gets \text{Val-Loss}_{\text{c}}$
    \ENDIF
\ENDFOR
\RETURN $\ma{m}^{*}_{\text{in}}, \ma{m}^{*}_{\text{arch}}$, Optimized model $\mathcal{F}^*$
\end{algorithmic}
\end{algorithm}

\subsection{Computational Complexity Reduction}
X-REFINE framework offers two computational complexity reduction levels, related to the model inputs and the architecture, respectively. The complexity is quantified by the floating-point operations (FLOPs) and expressed as follows:
\begin{equation}
\small
\begin{aligned}
\mathcal{C}(\ma{m}_{\text{in}}, \ma{m}_{\text{arch}}) &= \sum_{l=0}^{L-1} \left( 2 \cdot \|\ma{m}_{l}\|_0 \cdot \|\ma{m}_{l+1}\|_0 + \|\ma{m}_{l+1}\|_0 \right), \\
\text{where } \ma{m}_0 &= \ma{m}_{\text{in}}, \text{ and } \ma{m}_{\text{arch}} = [\ma{m}_1^T, \ma{m}_2^T, \dots, \ma{m}_{L-1}^T]^T.  
\end{aligned}
\end{equation}

$\ma{m}_l$ is a sub-mask representing the active neurons in the $l$-th hidden layer. $||\cdot||_0$ denotes the $\ell_0$-norm, which counts the number of active elements (non-zero entries) within each mask. The overall Complexity Reduction $ \Delta\mathcal{C}$ considering both the inputs and architecture optimizations is expressed as:

\begin{equation}
    \Delta\mathcal{C} = \left( 1 - \frac{\mathcal{C}(\ma{m}^{*}_{\text{in}}, \ma{m}^{*}_{\text{arch}})}{\mathcal{C}(\ma{m}^{0}_{\text{in}}, \ma{m}^{0}_{\text{arch}})} \right) \times 100\%.
\end{equation}

$\ma{m}^{0}_{\text{in}} = \mathbf{1}_{2K_{\text{on}}} \in \mathbb{R}^{2K_{\text{on}}}$ and $\ma{m}^{0}_{\text{arch}} = \mathbf{1}_{|\Theta|} \in \mathbb{R}^{|\Theta|}$ denote the original pre-trained model parameters with the full inputs and complete architecture.
\section{Simulation Results} \label{simulation_results}

The X-REFINE framework is evaluated using the STA-FNN estimator \cite{ref_survey} that employs a denoising FNN with three hidden layers to map the noisy $\hat{{\ma{h}}}_{\text{STA}_{q}}$ to a refined estimate $\hat{{\ma{h}}}_{\text{STA-FNN}_{q}}$. STA channel estimation is based on {\ac{DPA}} estimation followed by frequency and time averaging operations, as detailed in~\cite{ref_survey}.
The IEEE 802.11p standard is used with $K_{p} = 4$, $K_{d} = 48$, $K_{n} = 12$, and $I = 50$ OFDM symbols, in high-mobility ($f_d=1000$ Hz) VTV-EX low-frequency selective (LF) and VTV-SDWW high-frequency selective (HF) scenarios \cite{r19}. The model is trained on $10^5$ OFDM symbols (80\%/20\% split) using the Adam optimizer ($lr=10^{-3}$, 500 epochs, batch size 128). The analysis is performed according to five criteria\footnote{Throughout the BER figures' legends, for each XAI scheme we define (relevant FNN inputs, employed FNN architecture).}: (\textit{i}) X-REFINE optimization, evaluating both input and architecture optimization, (\textit{ii}) frequency selectivity, (\textit{iii}) modulation order, (\textit{iv}) interpretability resolution, where X-REFINE is benchmarked against XAI-CHEST, LIME, and DeepSHAP XAI methods, and (\textit{v}) computational complexity.

\begin{figure}[t]
\centering
\includegraphics[width=0.82\columnwidth]{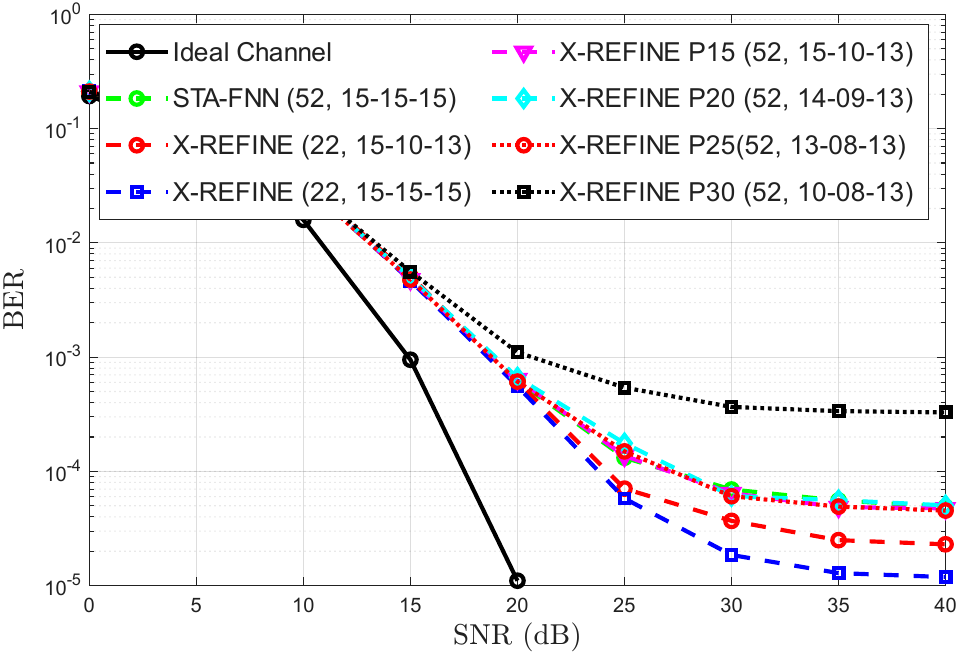}
\caption{Optimized performance of the proposed X-REFINE framework employing HF channel model and QPSK modulation order.}
\label{fig:proposed_opt_x_refine}
\end{figure}

\subsection{X-REFINE Double Optimization}

Threshold selection for input and architecture pruning is critical for optimizing BER performance, as formulated in (8). Leveraging the sign-stabilized LRP-$\epsilon$ rule, subcarriers are categorized based on relevance scores ($\bar{\ma{r}}$): (i) Reliable ($\bar{\ma{r}} \approx 1$): the subcarriers having the highest relevance scores. In the context of channel estimation, pilot subcarriers are considered reliable. (ii) Contributing ($\bar{\ma{r}} > 0$): the subcarriers having positive relevance scores are considered beneficial to the FNN-based channel estimation. (iii) Neutral ($\bar{\ma{r}} = 0$): zero relevance score signifies that the model is not considering the corresponding subcarriers at all in the channel estimation task. (iv) Harmful ($\bar{\ma{r}} < 0$),  negative relevance scores identify harmful subcarriers that actively mislead the FNN by propagating noise-induced errors, and thus degrade the FNN's estimation accuracy. We note that X-REFINE preserves only the contributing and reliable subcarriers as relevant inputs. For structural pruning, we evaluate percentile-based thresholds $P \in \{15, 20, 25, 30\}$. Recall that the objective here is to select the pruning percentile where the BER is minimized. Figure~{\ref{fig:proposed_opt_x_refine}} illustrates these optimizations for the HF channel model employing QPSK modulation. Notably, optimizing the FNN input while maintaining the full architecture, X-REFINE (Input), yields a more significant BER improvement than keeping the full input and optimizing only the architecture for different percentiles. This ensures that data-driven optimization, achieved by filtering relevant inputs, is more beneficial than model-driven optimization via architectural pruning. Finally, the double optimization achieves the best performance-complexity-interpretability trade-off, proving that high-quality input filtering is the primary driver of the performance gain.

\subsection{Impact of Frequency Selectivity}

Herein, the impact of frequency selectivity is investigated. Figure~3(a) shows the relevance score distribution considering the LF channel model. We can notice that the X-REFINE frameworks consider the pilots in the reliable zone regardless of the frequency selectivity and the modulation order. Moreover, the majority of the subcarriers are considered as neutral in the LF channel model due to the simplicity of the channel estimation task in such scenarios. In addition, considering only the pilots as FNN inputs provides the best BER performance in the LF channel model, as shown in Figures 4(a) and 4(b). This validates the importance of pilots taken by the FNN model. As the frequency selectivity becomes more challenging, the X-REFINE framework considers more contributing subcarriers. This is expected since in complicated scenarios, the FNN model needs more relevant inputs to accurately perform the estimation task. It is worth mentioning that the X-REFINE framework offers an architecture pruning by $P = 25\%$, from (15-15-15) to (14-11-09), and $P = 15\%$, from (15-15-15) to (14-13-12), in QPSK and 64QAM modulations, respectively. This signifies the need for higher structural capacity to maintain BER performance in higher modulation orders.


 \begin{figure*}[t]
	\centering	
    \includegraphics[width=1.89\columnwidth]{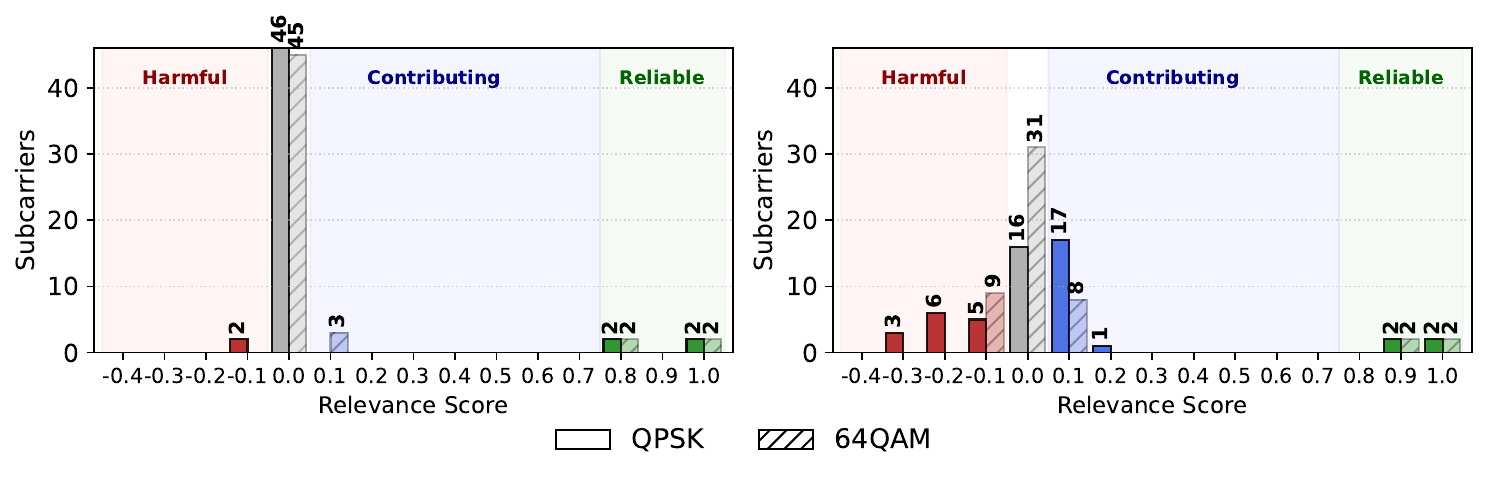}
	\caption{Relevance scores distribution of the proposed X-REFINE framework considering LF and HF channel models under different modulation orders: From left to right: (a) LF channel model, (b) HF channel model.}
	\label{fig:LF_BER}
\end{figure*}

\begin{figure*}[t]
	\centering	
    \includegraphics[width=2\columnwidth]{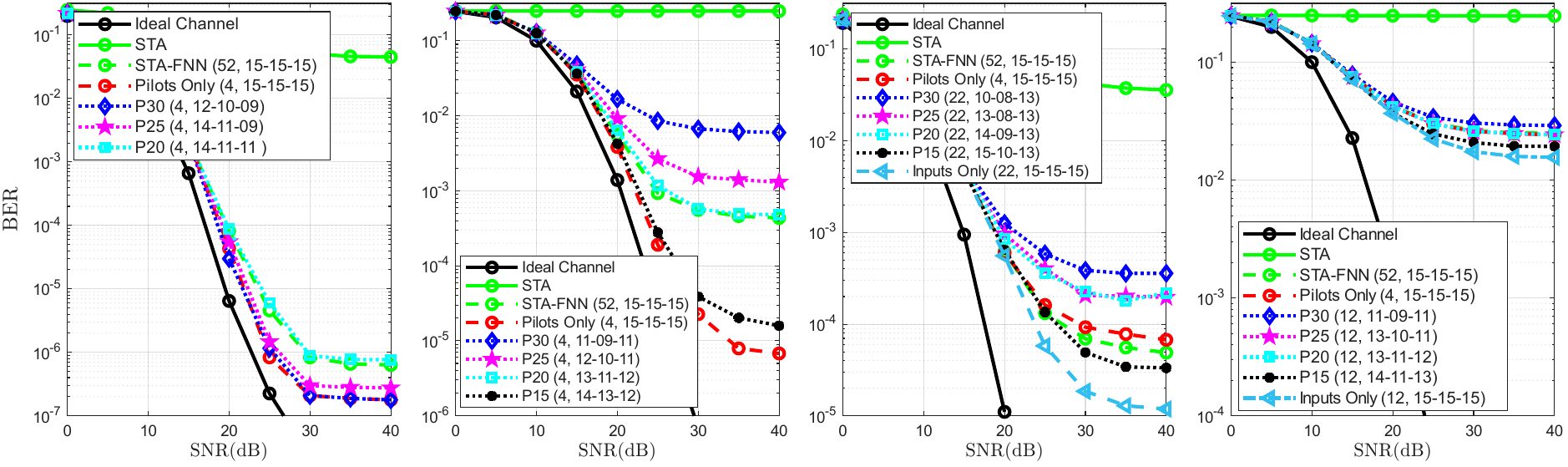}
	\caption{BER performance of the proposed X-REFINE framework for the LF and HF channel models under different modulation orders: From left to right: (a) LF - QPSK modulation, (b) LF - 64QAM modulation, (c) HF - QPSK modulation, and (d) HF - 64QAM modulation.}
	\label{fig:LF_BER}
\end{figure*}

\subsection{Impact of Modulation Order}

Figure~3(b) compares the relevance scores distributions of QPSK and 64QAM modulations, considering the HF channel model. We notice that for both modulations, the pilots are classified as reliable subcarriers, validating that the FNN recognizes pilots as indispensable for the channel estimation and independent of the employed modulation order. Moreover, as modulation order increases (QPSK to 64QAM) the number of neutral subcarriers nearly doubles (from 16 to 31). This indicates that in higher-order modulations, the FNN naturally neglects a larger portion of distorted noisy inputs, whereas in QPSK, the model attempts to process more subcarriers, resulting in a higher count of harmful inputs that propagate estimation errors. Therefore, as the modulation order increases, fewer relevant subcarriers are selected\footnote{Similar behavior trend is validated for 256QAM.}. Figures 4(c) and 4(d) illustrate the corresponding BER performance. X-REFINE selected $22$ and $12$ relevant subcarriers for QPSK and 64QAM modulations, respectively. Architectural pruning with $P=15\%$ provides the optimal performance-complexity trade-off. The first layer is consistently preserved to ensure reliable feature extraction, while pruning is concentrated in the middle layer, indicating that the FNN model prioritizes input integrity over excessive processing depth.

\subsection{Impact of Interpretability Resolution}

The interpretability resolution is defined as the capability of an XAI method to accurately identify truly relevant subcarriers. Figure 5 shows the relevance scores of several XAI methods employing the HF channel model. The proposed X-REFINE framework produces a significantly sparser distribution, effectively isolating contributing subcarriers from neutral ones. In contrast to benchmark methods,  LIME~\cite{ribeiro2016should}, DeepSHAP~\cite{lundberg2017unified}, and XAI-CHEST~\cite{gizzini2025explainable}, which rely on external processing, X-REFINE computes relevance scores by directly backpropagating the FNN output via the internal architecture. This structural alignment allows X-REFINE to achieve the best BER while selecting the minimal subset of subcarriers, as shown in Figure 6(a). This proves that internally based relevance attribution is more reliable than external approximations. Furthermore, the throughput analysis reveals that regardless of the input optimization applied, all models converge to the same maximum achievable throughput, Figures 6(b). This confirms that X-REFINE's double optimization significantly reduces computational overhead without compromising overall system efficiency. Consequently, the X-REFINE framework establishes a superior trade-off across performance, complexity, interpretability resolution, and system efficiency.

\subsection{Computational Complexity Analysis}

Table I summarizes the computational complexity reduction ($\Delta\mathcal{C}$) relative to the full, non-optimized FNN model, considering full inputs and full architecture. Moreover, it provides the inherent complexity of the XAI methods ($\mathcal{C}_{\text{XAI}}$) where $D_{\text{SHAP}}$ represents the number of background reference samples used in DeepSHAP, $D_{\text{LIME}}$ denotes the number of local perturbations generated per instance in LIME, and $E$ and $D_{\text{XAI-CHEST}}$ represent the training epochs and dataset size required to train the XAI-CHEST interpretability model. In the LF scenario, where only pilot subcarriers are retained as relevant inputs, reducing the FNN input size from 104 to 8 yields a baseline reduction of 39.51\%. X-REFINE further optimizes this by pruning the architecture, achieving total reductions of 62.41\% and 51.41\% for QPSK and 64QAM, respectively. While these gains are slightly lower in HF scenarios due to increased channel frequency selectivity, X-REFINE consistently outperforms the benchmarked XAI methods, offering up to 35.16\% and 43.59\% reduction for QPSK and 64QAM modulations, respectively. Beyond the optimized model complexity, the complexity of the XAI method is also critical. XAI-CHEST has a significant overhead since a training phase is required with ($\mathcal{O}(E D K_{\text{on}}^2)$) additional complexity. In contrast, X-REFINE is a training-free analytic framework with a minimal complexity of $\mathcal{O}(K_{\text{on}})$. By eliminating the training overhead, X-REFINE improves model optimization and provides a superior performance-complexity-interpretability trade-off. This makes it more adapted to latency-critical applications.

\begin{figure}[t]
    \centering
{\includegraphics[width=\columnwidth,height=4.5cm]{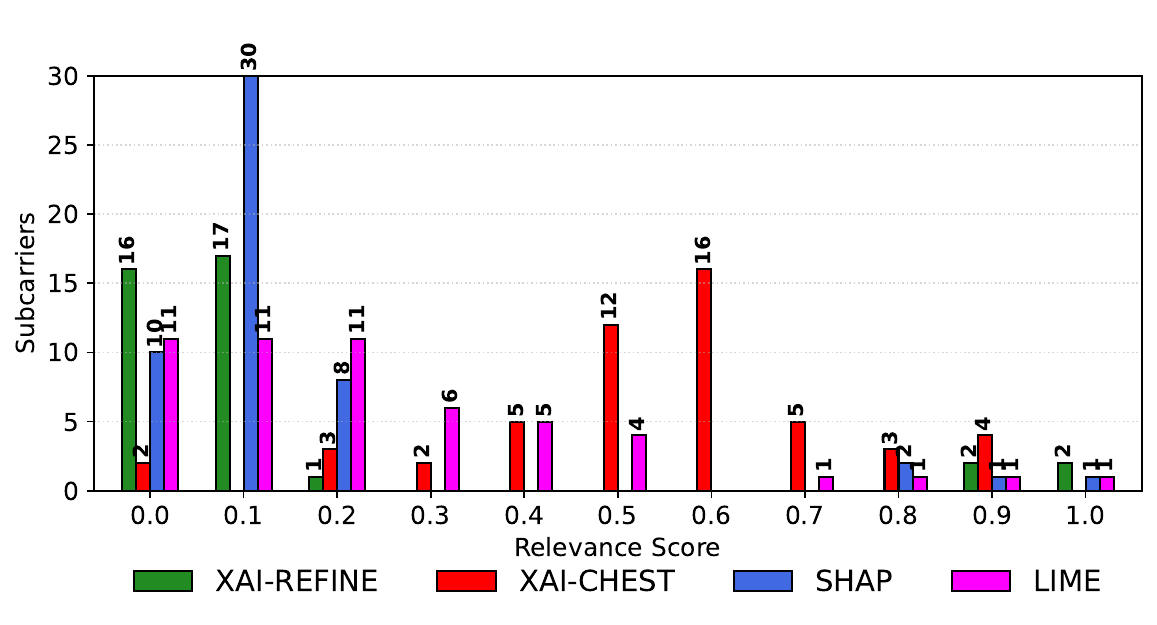}}%
    \caption{Relevance scores distribution of the benchmarked XAI schemes considering the HF channel model employing QPSK modulation order.}
    \label{fig:XAI_HF}
\end{figure}

\begin{figure}[t]
	\centering	    \includegraphics[width=\columnwidth]{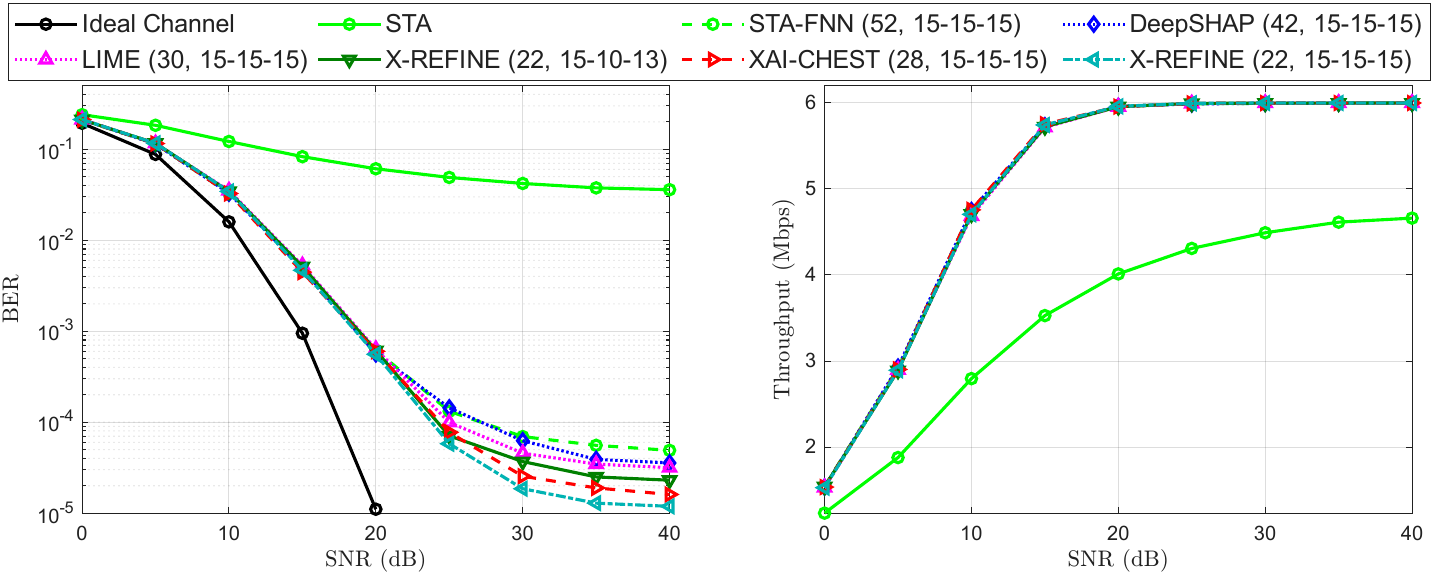}
	\caption{Performance evaluation of the benchmarked XAI methods for HF channel model employing QPSK modulation order: (a) BER, (b) Throughput.}
	\label{fig:XAI_BER_TH}
\end{figure}

\begin{table}[t]
\renewcommand{\arraystretch}{1.8}
\centering
\caption{Computational complexity comparison of $\mathcal{F}^{*}$ optimized by different XAI methods.}
\resizebox{\linewidth}{!}{
\begin{tabular}{|c|cccc|c|}
\hline
\multirow{3}{*}{\textbf{Method}} & \multicolumn{4}{c|}{\textbf{Complexity Reduction ($\Delta\mathcal{C}$)}}                                                                                       & \multirow{3}{*}{\textbf{$\mathcal{C}_{\text{XAI}}$}} \\ \cline{2-5}
                                 & \multicolumn{2}{c|}{LF}                                                       & \multicolumn{2}{c|}{HF}                                  &                                             \\ \cline{2-5}
                                 & \multicolumn{1}{c|}{QPSK}             & \multicolumn{1}{c|}{64QAM}            & \multicolumn{1}{c|}{QPSK}             & 64QAM            &                                            \\ \hline
LIME                             & \multicolumn{2}{c|}{\multirow{3}{*}{39.51\%}}                                 & \multicolumn{1}{c|}{18.11\%}          & 9.88\%           &  $\mathcal{O} (D_{\text{LIME}} K_{\text{on}}^2)$                                            \\ \cline{1-1} \cline{4-6} 
DeepSHAP                         & \multicolumn{2}{c|}{}                                                         & \multicolumn{1}{c|}{8.23\%}           & 14.82\%          &   $\mathcal{O} (D_{\text{SHAP}} K_{\text{on}})$                                          \\ \cline{1-1} \cline{4-6} 
XAI-CHEST                        & \multicolumn{2}{c|}{}                                                         & \multicolumn{1}{c|}{24.70\%}          & 32.93\%          &   \makecell{  $\mathcal{O} (E D_{\text{XAI-CHEST}} K_{\text{on}}^2)$ \\  +       $\mathcal{O} (K_{\text{on}})$}                  \\ \hline
\textbf{X-REFINE}                & \multicolumn{1}{c|}{\textbf{62.41\%}} & \multicolumn{1}{c|}{\textbf{51.41\%}} & \multicolumn{1}{c|}{\textbf{35.16\%}} & \textbf{43.59\%} & \textbf{$\mathcal{O} (K_{\text{on}})$}                                   \\ \hline
\end{tabular}
}
\end{table}

\section{Conclusion} \label{conclusions}

We propose X-REFINE, a decomposition-based XAI framework that leverages the LRP rule to optimize both the FNN model inputs and architecture simultaneously. Unlike perturbation-based XAI frameworks that identify only the relevant inputs based on external methodology, the proposed X-REFINE framework provides a double-pruning process based on the internal architecture of the model while maintaining a robust BER and reducing the overall computational complexity in different scenarios. Therefore, the proposed X-REFINE framework offers better performance-complexity-interpretability trade-offs and maintains the system efficiency, making it a robust solution for deploying high-performance, low-latency deep learning models in resource-constrained wireless communication systems. As future perspectives, we will focus on:  (\textit{i}) Adapting the X-REFINE framework to support recurrent neural networks (RNNs), and (\textit{ii}) Evaluating the performance of different gradient-based XAI schemes using the proposed X-REFINE framework.

{\small
\bibliographystyle{IEEEtran}
\bibliography{ref}

@ARTICLE{11435025,
  author={Zhang, Ping and Xu, Xiaodong and Sun, Mengying and Gao, Haixiao and Ma, Nan and Wang, Xiaoyun and Zhang, Ruichen and Wang, Jiacheng and Niyato, Dusit},
  journal={IEEE Communications Standards Magazine}, 
  title={{Toward Native AI in 6G Standardization: The Roadmap of Semantic Communication}}, 
  year={2026},
  volume={},
  number={},
  pages={1-11},
  doi={10.1109/MCOMSTD.2026.3669900}}

@inproceedings{lundberg2017unified,
author = {Lundberg, Scott M. and Lee, Su-In},
title = {{A Unified Approach to Interpreting Model Predictions}},
year = {2017},
isbn = {9781510860964},
publisher = {Curran Associates Inc.},
address = {Red Hook, NY, USA},
booktitle = {Proceedings of the 31st International Conference on Neural Information Processing Systems},
pages = {4768–4777},
numpages = {10},
location = {Long Beach, California, USA},
series = {NIPS'17}
}

@inproceedings{ribeiro2016should,
  title={{Why Should I Trust You? Explaining the Predictions of Any Classifier}},
  author={Ribeiro, Marco Tulio and Singh, Sameer and Guestrin, Carlos},
  booktitle={Proceedings of the 22nd ACM SIGKDD international conference on knowledge discovery and data mining},
  pages={1135--1144},
  year={2016}
}

@ARTICLE{10620685,
  author={Senevirathna, Thulitha and La, Vinh Hoa and Marcha, Samuel and Siniarski, Bartlomiej and Liyanage, Madhusanka and Wang, Shen},
  journal={IEEE Communications Surveys \& Tutorials}, 
  title={{A Survey on XAI for 5G and Beyond Security: Technical Aspects, Challenges and Research Directions}}, 
  year={2025},
  volume={27},
  number={2},
  pages={941-973},
  doi={10.1109/COMST.2024.3437248}}

@ARTICLE{10742571,
  author={Mekrache, Abdelkader and Mekki, Mohamed and Ksentini, Adlen and Brik, Bouziane and Verikoukis, Christos},
  journal={IEEE Communications Magazine}, 
  title={{On Combining XAI and LLMs for Trustworthy Zero-Touch Network and Service Management in 6G}}, 
  year={2025},
  volume={63},
  number={4},
  pages={154-160},
  doi={10.1109/MCOM.002.2400276}}

@ARTICLE{11217271,
  author={Jain, Kurunandan and Krishnan, Prabhakar and Pachiyannan, Prabu and Jaganathan, Logeshwaran and Khan, Muhammad Attique and Li, Yang},
  journal={IEEE Communications Standards Magazine}, 
  title={{Toward Smart 5G and 6G: Standardization of AI-Native Network Architectures and Semantic Communication Protocols}}, 
  year={2025},
  volume={},
  number={},
  pages={1-12},
  doi={10.1109/MCOMSTD.2025.3618114}}

@article{gizzini2025explainable,
  title={{Explainable AI for Enhancing Efficiency of DL-based Channel Estimation}},
  author={Gizzini, Abdul Karim and Medjahdi, Yahia and Ghandour, Ali J and Clavier, Laurent},
  journal={IEEE Transactions on Machine Learning in Communications and Networking},
  year={2025},
  publisher={IEEE}
}

@article{montavon2019layer,
  title={{Layer-Wise Relevance Propagation: An Overview}},
  author={Montavon, Gr{\'e}goire and Binder, Alexander and Lapuschkin, Sebastian and Samek, Wojciech and M{\"u}ller, Klaus-Robert},
  journal={Explainable AI: interpreting, explaining and visualizing deep learning},
  pages={193--209},
  year={2019},
  publisher={Springer}
}

@ARTICLE{11126933,
  author={Rodríguez-Piñeiro, José and Wei, Zhongxiang and Wang, Jingjing and Gutiérrez, Carlos A. and Correia, Luis M.},
  journal={IEEE Open Journal of Vehicular Technology}, 
  title={{6G-Enabled Vehicle-to-Everything Communications: Current Research Trends and Open Challenges}}, 
  year={2025},
  volume={6},
  number={},
  pages={2358-2391},
  doi={10.1109/OJVT.2025.3599570}}

@ARTICLE{9743925,
  author={Gizzini, Abdul Karim and Chafii, Marwa},
  journal={IEEE Access}, 
  title={{Low Complex Methods for Robust Channel Estimation in Doubly Dispersive Environments}}, 
  year={2022},
  volume={10},
  number={},
  pages={34321-34339},
  doi={10.1109/ACCESS.2022.3162928}}

@ARTICLE{10854503,
  author={Sun, Haochen and Liu, Yifan and Al-Tahmeesschi, Ahmed and Nag, Avishek and Soleimanpour, Mohadeseh and Canberk, Berk and Arslan, Hüseyin and Ahmadi, Hamed},
  journal={IEEE Open Journal of the Communications Society}, 
  title={{Advancing 6G: Survey for Explainable AI on Communications and Network Slicing}}, 
  year={2025},
  volume={6},
  number={},
  pages={1372-1412},
  doi={10.1109/OJCOMS.2025.3534626}}

@ARTICLE{ref_survey,
  author={Gizzini, Abdul Karim and Chafii, Marwa},
  journal={IEEE Access}, 
  title={{A Survey on Deep Learning Based Channel Estimation in Doubly Dispersive Environments}}, 
  year={2022},
  volume={10},
  number={},
  pages={70595-70619},
  doi={10.1109/ACCESS.2022.3188111}}

@ARTICLE{r19,
  author={I. {Sen} and D. W. {Matolak}},
  journal={IEEE Transactions on Intelligent Transportation Systems}, 
  title={{Vehicle–Vehicle Channel Models for the 5-GHz Band}}, 
  year={2008},
  volume={9},
  number={2},
  pages={235-245},}
}

\end{document}